\documentclass{article}
\usepackage{spconf,amsmath,graphicx}
\usepackage{amssymb,amsfonts}
\usepackage{cite}
\usepackage{booktabs}
\usepackage{url}
\usepackage{balance}
\usepackage{subfigure}
\usepackage{multirow}
\usepackage[normalem]{ulem}
\usepackage{booktabs}
\graphicspath{{./Figures/}}

\usepackage[dvipsnames]{xcolor}

\usepackage{tikz}
\usetikzlibrary{backgrounds}
\usetikzlibrary{arrows,shapes}
\usetikzlibrary{tikzmark}
\usetikzlibrary{calc}
\usepackage{blindtext}
\usepackage{tcolorbox}
\usepackage{tikz}
\usetikzlibrary{arrows,shapes,positioning,shadows,trees,mindmap}
\usepackage[edges]{forest}
\usetikzlibrary{arrows.meta}
\colorlet{linecol}{black!75}
\usepackage{xkcdcolors} 
\usepackage{tikz}
\usetikzlibrary{backgrounds}
\usetikzlibrary{arrows,shapes}
\usetikzlibrary{tikzmark}
\usetikzlibrary{calc}

\title{Enhancing Semantic Communication with Deep Generative Models \\ -- An ICASSP Special Session Overview}

\name{Eleonora Grassucci$^1$, Yuki Mitsufuji$^2$, Ping Zhang$^3$, and Danilo Comminiello$^1$ \thanks{Corresponding author's email: eleonora.grassucci@uniroma1.it.}}
\address{$^1$Sapienza University of Rome, Rome, Italy\\
$^2$Sony Group Corporation, Tokyo, Japan\\
$^3$Beijing University of Posts and Telecommunications, Beijing, China}

\begin{document}

\maketitle

\begin{abstract}
Semantic communication is poised to play a pivotal role in shaping the landscape of future AI-driven communication systems. Its challenge of extracting semantic information from the original complex content and regenerating semantically consistent data at the receiver, possibly being robust to channel corruptions, can be addressed with deep generative models.
This ICASSP special session overview paper discloses the semantic communication challenges from the machine learning perspective and unveils how deep generative models will significantly enhance semantic communication frameworks in dealing with real-world complex data, extracting and exploiting semantic information, and being robust to channel corruptions. Alongside establishing this emerging field, this paper charts novel research pathways for the next generative semantic communication frameworks. 
\end{abstract}


\section{Introduction}

Generative semantic communication is an emerging topic that merges the fresh semantic communication area and the cutting-edge field of deep generative modeling. 
The next 6G communication systems will be AI-based and will rely on the transmission of the semantic information, trying to regenerate a semantically equivalent content, rather than exactly recovering the original bit sequence \cite{Qin2021SemanticCP, Barbarossa2023COMMAG, Dai2021CommunicationBT, Huang2021GLOBECOM}. 
Semantic communication allows greater transmission efficiency due to the extraction and compression of semantic information leveraging effective deep learning models. As a consequence, the transmitted semantics are generally smaller and more robust to communication distortions, allowing to reduce latency and bandwidth while preserving reliability. This new paradigm is gaining interest in several fields of application, ranging from image compression \cite{Patwa2020SemanticPreservingIC}, to video transmission \cite{Jiang2022WirelessSC}, speech \cite{Xiao2023ICASSP, Han2022SemanticPreservedCS, Weng2021ICC, Weng2023TWC}, point clouds \cite{han2023semanticaware}, and the metaverse \cite{Park2022EnablingTW, Jagatheesaperumal2023SemanticawareDT}.
With the increasing amount of connected devices and produced multimedia content, next-generation communication systems open new challenges. In the near future, we expect an explosion of demands that will require high compression ratios of transmitted data, leading to the study of how to properly extract semantic data, how to handle and exploit the received information or make receivers robust to channel corruptions and distortions.

In this scenario, deep generative models assume crucial importance due to their ability to extract and exploit the received semantic information through a semantic-guided generation process. Such models are trained to learn the original data distribution with the aim of generating new points by sampling from it. Conditioning this generation process allows the user to control the output of the generation and generate content starting from a textual prompt, a low-resolution image, or a low-dimensional map. For these reasons, generative models can significantly enhance semantic communication frameworks, solving most of the learning challenges this new paradigm sets.
Very recently, a few examples have been proposed, including generative adversarial networks (GANs)~\cite{goodfellow2014generative, takida2023san} that have been involved to generate content from the received semantic vector \cite{Han2022GenHighEff, Gunduz2022GenSem, Han2022SemanticPreservedCS}, or from the semantic map \cite{Huang2021GLOBECOM}. Similarly, variational autoencoders (VAEs)~\cite{kingma2013auto, takida2022sq-vae} have been exploited for joint source-channel coding and image transmission \cite{Nemati2023, Malur2020VAE, Estiri2020AVA}. More recently, the power of state-of-the-art diffusion models~\cite{ho2020denoising, lai2023fp} has been exploited to generate photorealistic cityscape images starting from the corrupted received semantic maps with a very robust framework \cite{Grassucci2023GenerativeSC} or from a textual prompt to reduce energy consumption \cite{Lee2023EnergyEfficientDS}.
Together with their promising solutions, generative semantic communication frameworks open new challenges and research pathways concerning the definition of semantic communication-tailored training losses and generative models. In particular, the sustainability of large state-of-the-art generative models will be crucial, together with novel methods to exploit the received noisy semantics and the assessment of the quality of service.

While several overviews exist on semantic communication, they usually face the topic from a communication perspective. On the contrary, the aim of this ICASSP special session overview paper is to present a novel viewpoint on semantic communication challenges specific from the machine-learning point of view. Moreover, this work presents a novel solution to such challenges, showing how generative models can remarkably enhance semantic communications, and tabling future research directions for this emerging topic.
Accordingly, the rest of the paper is organized as follows. Section \ref{sec:challenges} describes the machine learning challenges in semantic communications, Section \ref{sec:generative} presents deep generative modeling and its opportunities in semantic communication drawing future directions, while Section \ref{sec:con} concludes the paper.


\section{Machine learning challenges in Semantic communication}
\label{sec:challenges}

Semantic communication lays its foundations in the three communication levels theorized by Weaver \cite{Weaver1953levels}:
\begin{itemize}
    \item Level A: The technical level. It focuses on the most accurate way of transmitting bits from the sender to the receiver.
    \item Level B: The semantic level. It focuses on the best way of transmitting the meaning of the messages rather than accurately recovering the transmitted bits.
    \item Level C: The effectiveness level. It focuses on the efficiency of the transmission of previous levels.
\end{itemize}

\begin{figure}
    \centering
    \includegraphics[width=\linewidth]{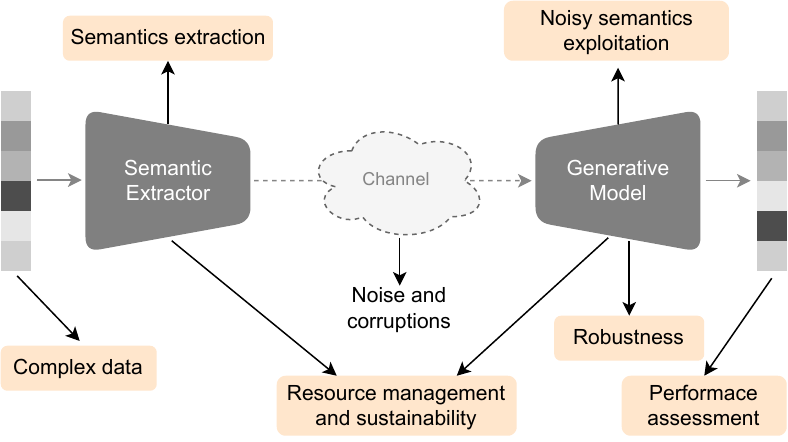}
    \caption{Semantic communication framework overview with its challenges from the machine learning perspective.}
    \label{fig:framework}
\end{figure}

As existing communication systems have nearly approached the physical-layer capacity limit theorized by Shannon \cite{shannon1948mathematical}, a radical rethinking of the communication technologies design has been promoted, moving from Level A to Level B of Weaver's theory. Semantic communication has been presented as the core of the next 6G communications that will be AI-driven \cite{Strinati20206GNB}. It relies on the transmission of the content semantic features, i.e., the meaning of the content, rather than on the accurate bitstream recovery. Such meaning can also weave together the goal of communication and the task the receiver has to solve. 
This change of paradigm opens new challenges both from the communication and the machine learning points of view. Several works already investigate communication challenges, while we focus on the unexplored machine learning perspective since AI-based semantic communication frameworks have to deal with the following issues. Figure~\ref{fig:framework} presents a visual representation of a semantic communication framework with corresponding machine learning challenges.

\textbf{Complex data.} Transmitted data is becoming much more complex and multimodal, as the number of connected multimedia devices is growing and permeating our daily lives. Applications such as video conferencing, metaverse, and autonomous driving require a large model capability both to extract and process such complex information. These domains need powerful frameworks able to model the compounded relations in 2D as well as in 3D data.

\textbf{Semantics extraction.} As the transmitted data becomes more complex and composed of multiple modalities such as videos comprising frames and audio, various sensors in IoT devices, and so on, the process of semantic extraction may not have a unique recipe. Indeed, the semantics may strongly depend on the type of data and on the task the receiver has to solve with this information. For example, the semantic map is a good representation of image semantics, but it does not exist for an audio signal that may need another semantic extraction. Equivalently, if the receiver task is to recognize the position of a pedestrian on the street, the textual description of \textit{``a pedestrian on the street"} may be insufficient, while the semantic map or the bounding boxes provide more insightful information.

\textbf{Noisy semantics exploitation.} Elaborating the received information in a lower-dimensional space and returning it to the data domain is a tricky task, whose difficulty increases as the data complexity grows. In addition, the received information may be corrupted by the channel, as we delineate in the next paragraph.

\textbf{Robustness.} The information transmitted over the communication channel may be corrupted by the noise or undergo transformations that deteriorate it. The degradation may include additional noise and loss of information such as pixels in images, frames in video, and so on. The receiver should be able to model these kinds of corruptions and exploit the received information.

\textbf{Performance assessment.} As the scope of the semantic communication is not the exact and accurate bit recovery at the receiver side, standard metrics such as MSE, SSIM, and PSNR are not appropriate anymore. Indeed, these metrics evaluate the reconstruction accuracy of the content, where each part of the signal is equally considered. On the contrary, semantic communication frameworks try to optimize the semantics of the message and the performance assessment should take the semantics into account. 

\textbf{Resource management and sustainability.} Both the sender and the receiver networks are subjected to resource constraints due to different aspects: i) They may have hardware limitations while current machine learning models have high computational demands; ii) The sender should reduce as much as possible the bandwidth usage and compress the semantic information. 



\section{Deep Generative Models and Semantic Communication}
\label{sec:generative}



\subsection{Deep Generative Models}

State-of-the-art deep generative models have demonstrated impressive results in dealing with real-world data of different types, ranging from very complex images \cite{Rombach2021latent, saharia2022photorealistic} to singing voice \cite{Takahashi2023ICASSP} and any kind of audio \cite{Ghosal2023TexttoAudioGU}, 3D point clouds \cite{kong2023generative} and structures \cite{luo2021a}, or video \cite{hong2023cogvideo, Singer2022MakeAVideoTG}. Generative deep learning models learn to model the data distribution during training. In the generation phase, they sample from the learned distribution new generated samples. This formulation allows sufficiently sophisticated generative models to learn any data distribution, regardless of its complexity.

While learning the data distribution, many generative models exploit representation learning to build or make use of lower-dimensional latent spaces. Often, while modeling this latent space or learning the data distribution, generative models build a semantic representation of the data. Variational autoencoders (VAEs) extract semantic features from data with a recognition model and naturally shape a latent space that can be regularized to place similar images close in the semantic space \cite{xie2021acvae}. Similarly, Generative adversarial networks (GANs) showed the ability to encode semantics in their feature maps in a linearly separable form \cite{Xu2021LinearSI, xu2022extracting}. Moreover, diffusion models intrinsically have a semantic space in which a small shift reflects a small change in the corresponding attribute of the generated image, without needing to retrain the network \cite{kwon2023diffusion}. In addition, intermediate activations of the diffusion model network that performs the reverse process indeed capture the semantic information of the input image, building pixel-level representations of the data \cite{baranchuk2022labelefficient}. However, contrary to VAEs and GANS, diffusion models semantic space is not lower-dimensional since it necessarily has to be equal to the data dimensionality.

Moreover, deep generative models perfectly fit the semantic communication scenario due to their ability to exploit semantic information. Indeed, there exist several state-of-the-art methods that guide the generation by means of different kinds of semantics. An example is semantic image synthesis in which the generative model is conditioned by the semantic map and generates images according to this information \cite{Wang2022SemanticIS}. Another explanatory case is text-to-speech synthesis, in which the semantic is the textual transcription of the speech the generative model will synthesize \cite{Kaur2023AIReview}. 
Formally, a generic formulation for a semantic-guided deep generative model can be the following. Assuming that $\mathbf{s}$ is the received semantics and $\mathcal{D}$ the data domain, the generative semantic communication framework can be formalized as
\begin{equation}
    \mathbf{x} = G(\mathbf{\epsilon} , \mathbf{s}),
\end{equation}
\noindent with

\begin{equation}
    \mathbf{x} \in \mathcal{D}, \qquad \mathbf{\epsilon} \sim \mathcal{N}(\mathbf{0}, \mathbf{I}).
\end{equation}

\noindent The solution consists of learning the generative model $G$ that transforms the realization $\mathbf{\epsilon}$ of a distribution from which it is easy to sample such as the standard Gaussian into the data $\mathbf{x}$ under the conditioning guidance of the semantics $\mathbf{s}$.

\subsection{Generative Semantic Communication: Opportunities and Future Directions}

In this Subsection, we delineate possible pathways to cover in generative semantic communication and the opportunities this emerging topic introduces.

First and most importantly, novel \textbf{generative semantic communication frameworks} should be developed. Existing state-of-the-art deep generative models are not tailored for communication systems and, to be incorporated/exploited for this task, they have to be re-engineered and inserted in the right framework. Indeed, state-of-the-art models are not trained with noisy data, while directly instructing such models by simulating channel noise in the training data may result in more robust generative models \cite{Grassucci2023GenerativeSC}.
Similarly, a challenge generative semantic communication frameworks have to face is exploiting noisy semantic information since state-of-the-art deep generative models consider clean conditioning. Guiding the generation with noisy conditioning may inject distorted information into the sampling process and lead to out-of-distrbution or, in turn, to noisy and corrupted generated samples. 

Novel \textbf{semantic extraction and exploitation approaches} may be developed. As an example, text embeddings, although sometimes less accurate, may reduce communication costs with respect to semantic maps for several objects. Building a trade-off between the accuracy of the semantic information, the transmission cost/bandwidth, and the goal of the receiver is a challenging task that novel generative semantic communication frameworks have to face. In addition, solutions comprising hybrid encoders for semantics according to the importance of semantic information portions are very promising \cite{Dai2021CommunicationBT, du2023yolobased}.

Another possible direction for the next generative semantic communication frameworks is defining novel \textbf{loss functions} that consider the preservation of the semantics in the regenerated content rather than the bitstream recovery. Indeed, classical machine learning communication systems are trained so as to maximize the transmission accuracy, as measured by reconstruction and recovery loss. Changing perspective from classical to semantic communication systems means that the quality of service should be measured by losses tailored to the meaning preservation and the receiver task. Concurrently, novel performance assessment strategies have to be developed, specifically customized for evaluating the different types of transmitted semantics and their preservation in multimedia regenerated content.

One of the core opportunities for generative semantic communication research lies in the development of \textbf{sustainable and resource-efficient frameworks}. The weaved-together opportunities here are multiple. First, deep generative models work their magic at the cost of billions of learnable parameters that translate into large storage memory demands. As a quantitative example, the first version of Stable Diffusion \cite{Rombach2021latent}, which is one of the most lightweight among the state-of-the-art text-to-image models, has around 890M parameters that translate into approximately 3.5GB of storage memory on the device. A step towards reducing the storage memory of generative models has been covered with the definition of these networks in the quaternion domain that allows a reduction up to the $75\%$ of the memory while obtaining comparable generation abilities \cite{GrassucciQGAN2021, grassucci2021lightweight}. Second, several strategies have to be found to reduce the long inference time of state-of-the-art diffusion models in order to work in online scenarios. Third, such large-scale sizes and the long time of training and inference obviously lead to large requirements of computational resources and consequently very high energy consumption. The training of such huge models necessarily needs GPUs with large VRAM and often takes days of computation, producing a massive amount of CO2. Moreover, once the training is complete, the inference phase usually requires GPUs too, even though with lower requirements and energy consumption with respect to training, that are hard to embed on smaller devices.
Therefore, in the next generative semantic communication frameworks resource optimization will assume a crucial role \cite{Karapantelakis2023Annals}. New techniques can be developed to reduce the computational requirements or make network blocks reusable by considering pruning, quantization, and modular networks. Such methods may help make generative semantic communications frameworks greener and embeddable in smaller devices.

Additionally, to improve sustainability, new frameworks should begin to release code and pretrained weights to make them accessible and available for future applications and researchers without forcing the latter to perform new training and consume more energy. Such pretrained models can serve as a basis for the next frameworks and they can be fine-tuned on specific tasks or different datasets from other users, saving training energy and consistently reducing the amount of CO2 produced.



\section{Conclusion}
\label{sec:con}

In this ICASSP special session overview paper, we propose a novel perspective on semantic communications, which will be the core of future 6G communications. We analyze the challenges that machine learning models have to face in semantic communication frameworks. We highlight how deep generative models can significantly enhance the next semantic communication frameworks and we provide new research opportunities and directions for the generative semantic communication research topic. Addressing these challenges will help the development of future AI-driven semantic communication systems.

\balance
\ninept
\bibliographystyle{IEEEbib}
\bibliography{Biblio}

\end{document}